# Learning Contextualised Cross-lingual Word Embeddings and Alignments for Extremely Low-Resource Languages Using Parallel Corpora

**Takashi Wada**[1*] **Tomoharu Iwata**[2,3] **Yuji Matsumoto**[3*] **Timothy Baldwin**[1] **Jey Han Lau**[1]

[1] School of Computing and Information Systems, The University of Melbourne
[2] NTT Communication Science Laboratories
[3] RIKEN Center for Advanced Intelligence Project (AIP)
twada@student.unimelb.edu.au,
tomoharu.iwata.gy@hco.ntt.co.jp, yuji.matsumoto@riken.jp,
{tbaldwin, jeyhan.lau}@unimelb.edu.au

## Abstract

We propose a new approach for learning contextualised cross-lingual word embeddings based on a small parallel corpus (e.g. a few hundred sentence pairs). Our method obtains word embeddings via an LSTM encoder-decoder model that simultaneously translates and reconstructs an input sentence. Through sharing model parameters among different languages, our model jointly trains the word embeddings in a common cross-lingual space. We also propose to combine word and subword embeddings to make use of orthographic similarities across different languages. We base our experiments on real-world data from endangered languages, namely Yongning Na, Shipibo-Konibo, and Griko. Our experiments on bilingual lexicon induction and word alignment tasks show that our model outperforms existing methods by a large margin for most language pairs. These results demonstrate that, contrary to common belief, an encoder-decoder translation model is beneficial for learning cross-lingual representations even in extremely low-resource conditions. Furthermore, our model also works well on high-resource conditions, achieving state-of-the-art performance on a German-English word-alignment task.[1]

## 1 Introduction

Cross-lingual word embedding learning has the goal of learning representations for words of different languages in a common space (Mikolov et al., 2013b; Conneau et al., 2018; Levy et al., 2017). Cross-lingual representations are beneficial for finding correspondences between languages, and are utilised in many downstream tasks such as machine translation (Lample et al., 2018; Artetxe et al., 2018b) and cross-lingual named entity recognition (Xie et al., 2018).

The recent trend in cross-lingual embedding models is to leverage an enormous amount of monolingual data for each of the target languages, e.g. by training word embeddings monolingually and mapping them into a common space. Another approach is to jointly train cross-lingual word embeddings in the same space. Recently, cross-lingual masked language models such as mBERT (Devlin et al., 2019) have succeeded in learning cross-lingual representations using large-scale monolingual data for multiple languages.

However, when dealing with endangered languages, there is generally no such large-scale corpus of monolingual data, and when trained under low-resource settings, modern pretraining methods do not perform well (Hu et al., 2020). In this paper, we propose a joint-training method that learns contextualised cross-lingual word embeddings using a small parallel corpus, of the scale and form constructed by field linguists for language documentation purposes. Compared to previous models based on parallel corpora, our model has two strengths: (1) while previous models extend bag-of-words models such as Skip-Gram (Mikolov et al., 2013a) and capture only rudimentary word order information, our model encodes sentences with LSTMs (Hochreiter and Schmidhuber, 1997) and generates contextualised word embeddings; and (2) our model trains subword-aware word embeddings and captures orthographic similarities among the languages.

We perform evaluation over bilingual lexicon induction and word alignment. Both tasks are extremely important to facilitate language documentation, revitalisation, and education of endangered languages. We run experiments targeting three endangered languages — Yongning Na, Shipibo-Konibo, and Griko (Section 3.1) — as well as four high-resource language pairs, and show that our model substantially outperforms strong baselines for most language pairs.

---

[*]This work was partially done at Nara Institute of Science and Technology.

[1]Our code is available at `https://github.com/twadada/multilingual-nlm`

## 2 Methodology

### 2.1 Model Architecture

Our proposed model is based an LSTM[2] encoder-decoder model with attention (Luong et al., 2015b), trained with translation and reconstruction objectives (Figure 1). Suppose our model encodes a sentence $\langle x_1^s..., x_N^s \rangle$ in the source language $s$ and decodes a sentence $\langle y_1^t..., y_M^t \rangle$ in the target language $t$. The encoder employs bi-directional LSTMs $f$, which are shared among all languages:

$$r_i^s = E^s x_i^s, \qquad (1)$$
$$u_1^s..., u_N^s = f(r_1^s..., r_N^s), \qquad (2)$$

where $x_i^s$ denotes a one-hot vector. In cross-lingual tasks, we employ $r_i^s$ and $u_i^s$ as the static and contextualised word embeddings of $x_i^s$. Given the encoder states $u^s$, the decoders $\overrightarrow{g}^t$ and $\overleftarrow{g}^t$ translate (when $s \neq t$) or reconstruct (when $s = t$) the input sentence left-to-right and right-to-left. We train separate decoders for each language and direction to allow for the differences of word order.[3] Similar to ELMo (Peters et al., 2018), the decoding is performed independently in both directions:

$$r_i^t = E^t y_i^t, \qquad (3)$$
$$p(y_1^t..., y_M^t, \text{EOS}) = \prod_{i=1}^{M+1} p(y_i^t | \overrightarrow{h}_i, u^s),$$
$$\overrightarrow{h}_i = \overrightarrow{g}^t(\overrightarrow{h}_{i-1}, r_{i-1}^t),$$
$$p(\text{BOS}, y_1^t..., y_M^t) = \prod_{i=0}^{M} p(y_{M-i}^t | \overleftarrow{h}_{M-i}, u^s),$$
$$\overleftarrow{h}_i = \overleftarrow{g}^t(\overleftarrow{h}_{i+1}, r_{i+1}^t).$$

The output layer and attention mechanism are shared across the two directions:

$$p(y_i^t | h_i, u^s, r^s) = \text{softmax}(E^{t\top} h_i'), \qquad (4)$$
$$h_i' = W(\bar{u}_i^s + \bar{r}_i^s + h_i), \qquad (5)$$
$$\bar{u}_i^s = \sum_{j=1}^{N} \alpha_{i,j}^t u_j^s, \quad \bar{r}_i^s = \sum_{j=1}^{N} \alpha_{i,j}^t r_j^s, \qquad (6)$$
$$\alpha_{i,j}^t = \frac{\exp(h_i u_j^s)}{\sum_{k=1}^{N} \exp(h_i u_k^s)}, \qquad (7)$$

---

[2]We use LSTM rather than Transformer (Vaswani et al., 2017) because LSTM is less sensitive to hyper-parameters and performs better at translation under extremely low-resource conditions (Zhang et al., 2020).

[3]In our preliminary experiments, we have found that learning language-specific decoders improves cross-lingual embeddings for distant languages, e.g. SOV and SVO languages.

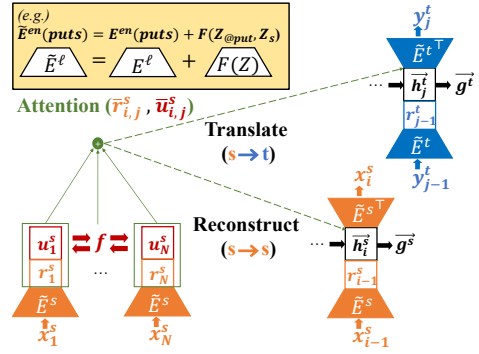

Figure 1: Our proposed model.

where $h_i$ denotes either $\overrightarrow{h}_i$ or $\overleftarrow{h}_i$, and $N$ is the number of words in the source sentence $x^s$. In Eqn. (4), we use the word embedding parameters $E^t$ for the output layer (weight tying: Inan et al. (2017); Press and Wolf (2017)). This technique can reduce the number of the language-specific parameters substantially, encouraging the model to use the same space across languages. When calculating attention weight in Eqn. (7), the model uses dot products of the encoder and decoder hidden states to encourage them to be in the same embedding space. In Eqn. (6), our model attends to the word embeddings as well as the hidden states to consider more direct relations between source and target word embeddings, i.e. $r_i^{t\top} W \bar{r}_i^s$ directly contributes to the probability $p(y_i^t)$.[4] Furthermore, we employ very aggressive dropout (Srivastava et al., 2014), which is applied to all the input and output word embeddings $E^\ell$ in Eqns. (1), (3), and (4), as well as to $\bar{u}_i^s + \bar{r}_i^s + h_i$ in Eqn. (5) before the linear transformation, with the dropout rate all set to 0.5. We show that this strong regularisation leads to better cross-lingual representations.

### 2.2 Shared Subword Embeddings

To incorporate orthographic information into word embeddings, we propose a simple yet effective method to combine word and subword embeddings, inspired by FastText (Bojanowski et al., 2017). For each word $w_i^\ell$, we calculate its subword-aware word embedding $\tilde{E}_{w_i}^\ell$ as follows:

$$\tilde{E}_{w_i}^\ell = E_{w_i}^\ell + \text{F}(Z_{k \in \text{Q}(w_i)}), \qquad (8)$$

---

[4]This simplifies the lexical model proposed by Nguyen and Chiang (2018). While they apply separate output layers to the weighted average of word embeddings and hidden states, we have found that sharing the same output layer ($E^t$) performed the best, suggesting that the optimal model architecture is different between word and sentence translations.

where F(·) denotes the subword encoding function; $Z_k$ denotes the $k$-th subword embedding and $Q(w_i)$ denotes the indices of the subwords included in $w_i$. The subword embeddings $Z$ are shared among all languages, capturing orthographic similarities across languages. For the encoding function F(·), we experiment with two methods:[5] (1) average pooling ("**SW<sub>ave</sub>**"); and (2) applying a convolutional neural network (CNN) function which is shared among all languages, followed by average pooling ("**SW<sub>cnn</sub>**"). For instance, the embedding of the English word *puts* is represented by its language-specific word embedding $E^{en}_{puts}$ and shared subword embeddings $Z_{@put}$ and $Z_s$, where @ in *@put* denotes the beginning of a word. To segment words into subwords, we apply SentencePiece (Kudo and Richardson, 2018).[6]

Distinct from a standard NMT model, we use the subword-aware embeddings $\tilde{E}^\ell$ not only as the input layers in Eqns. (1) and (3), but also as the output layer in Eqn. (4) (Figure 1). In this way, we encourage the model to learn subword correspondences between the source and target languages through attention, i.e. Eqns. (4) and (5). In monolingual language modelling, Assylbekov and Takhanov (2018) have previously shown the effectiveness of sharing morpheme-aware embeddings between the input and output layers.

## 2.3 Training

Given a parallel corpus aligned between languages $s$ and $t$, our model is trained to minimise the following training loss $J = J_{s,t} + J_{t,s}$, where $J_{s,t} = \sum_{j=1}^{C_{s,t}} \Delta(x_j, p(x^s_j|x^s_j)) + \Delta(y_j, p(y^t_j|x^s_j))$. Here, $C_{s,t}$ is the number of aligned sentences between languages $s$ and $t$, and $\Delta$ denotes the cross entropy loss. The first and second terms represent the reconstruction and translation loss, respectively. Our model can also take multiple parallel corpora as input and generate multilingual word embeddings. In this case, we sum the loss calculated on each parallel corpus. For instance, we train our multilingual model to minimise the loss $J = J_{nru,en} + J_{en,nru} + J_{nru,fr} + J_{fr,nru} + J_{nru,zh} + J_{zh,nru}$ given three parallel corpora of nru–en, nru–fr, and nru–zh,[7] where some sentences are aligned between more than two languages. To balance the corpus size, we perform oversampling to ensure that $C_{s,t}$ is the same regardless of the language pairs $s$ and $t$. Our model hyper-parameters are detailed in Appendix A.[8]

---

[5] Dropout is always applied to subword embeddings.
[6] We also tried using character n-grams like FastText, but observed worse results. For Chinese, we segment words into characters due to the large number of character types.
[7] "nru" is the ISO-639-3 language code for Na.

| src–tgt | #Sents | #Tokens src | #Tokens tgt | #Vocab src | #Vocab tgt |
|---|---|---|---|---|---|
| nru-en | 605 | 4,690 | 7,849 | 1,860 | 942 |
| nru-fr | 3,833 | 33,816 | 75,997 | 8,801 | 4,797 |
| nru-zh | 1,766 | 15,127 | 21,598 | 4,704 | 2,459 |
| shp-es | 14,276 | 198,024 | 214,127 | 20,654 | 14,781 |
| grk-it (s) | 330 | 2,374 | 2,384 | 689 | 456 |
| grk-it (l) | 9,788 | 207,294 | 178,980 | 12,697 | 9,813 |

Table 1: The statistics of the parallel corpora of endangered languages used in this paper. grk-it (s) and (l) denote the smaller and larger corpora, respectively.

## 3 Experiments

### 3.1 Data

We conduct experiments on real-world data sets for three endangered languages: Yongning Na (nru), Shipibo-Konibo (shp), and Griko (grk) (Table 1).[9] The URLs of the data are shown in Appendix B.

#### 3.1.1 Yongning Na

Yongning Na is an unwritten Sino-Tibetan language with less than 50,000 speakers (Do et al., 2014). Due to the lack of a writing system, textual data must be professionally transcribed from speech by linguists, which precludes the use of the latest pretraining methods such as BERT (Devlin et al., 2019). As a cross-lingual resource for Na, there exists a phonemically transcribed corpus that has been translated into French, Chinese, and English, which is a part of the Pangloss Collection (Michailovsky et al., 2014). However, there are two challenges in learning cross-lingual representations from this data. First, the syntax and orthography of the languages are very different: Na is an SOV language with rich tonal morphology (Michaud, 2017), while the others are SVO languages. Second, there is a lot of noise in the parallel corpora. For instance, some words or phrases in the translations are written in brackets, indicating alternative translations, subsidiary information, or words that are implicit in the original Na sentences (Table 2). To clean the data, we use the pre-

---

[8] They are tuned on a small subset of de-en or fr-en data.
[9] Note that Griko is not included in ISO-639-3, and "grk" is an arbitrary (non-assigned) designator used in this paper.

| Na | raw | njɤ˩ ǀ ɑɹo˩ ǀ ə˧si˧-ɳɯ˧ ʐwɤ˩qʰv˩mv˩-hĩ˩ lɑ˩ ɲi˩ mæ˩! |
|  | cln | njɤ˩ ɑɹo˩ ə˧si˧-ɳɯ˧ ʐwɤ˩qʰv˩mv˩-hĩ˩ lɑ˩ ɲi˩ mæ˩ |
| English | raw | It (i.e. this story) is only what (we've) heard our great-grandmother tell. |
|  | cln | it is only what heard our great grandmother tell |

Table 2: An example Na–English parallel sentence before and after pre-processing ("raw" vs. "cln")

processing code used in Adams et al. (2017) with minor modifications.[10]

### 3.1.2 Shipibo-Konibo

Shipibo-Konibo is an indigenous language spoken by around 35,000 native speakers in the Amazon region of Peru (Vasquez et al., 2018), and is "definitely endangered" according to the UNESCO's Atlas of the World's Languages in Danger (Moseley, 2010). There is no large monolingual corpus for the language,[11] but for cross-lingual resources there are two parallel corpora aligned with Spanish, which are extracted from the Bible and educational books (Galarreta et al., 2017). Similar to Na, Shipibo-Konibo is an SOV language with very rich morphology (Valenzuela, 1997; Vasquez et al., 2018), whereas Spanish is an SVO language.

### 3.1.3 Griko

Griko is a Greek dialect spoken in southern Italy, and "severely endangered" according to UNESCO. There is no large-scale monolingual corpus of Griko, but there are two Griko–Italian parallel corpora (Zanon Boito et al., 2018; Anastasopoulos et al., 2018), with the smaller one including gold word alignment annotations. However, Griko has never had a consistent orthography, and hence its tokenisation and word segmentation differ across these corpora: the smaller data set is based on orthographic conventions from Italian, while the larger one follows the concept of a phonological word (Anastasopoulos et al., 2018). Unlike the Na and Shipibo-Konibo data sets, Griko and Italian are very similar in many ways: they both use the Latin script and have similar syntax. Therefore, the main challenge comes from the data paucity and inconsistent orthography in Griko, both of which are common problems for endangered languages.

### 3.2 Baselines

We compare our model against various cross-lingual models that are trained on a parallel corpus. First, we compare our model against a recently-proposed word-alignment model based on **mBERT** (Dou and Neubig, 2021).[12] It fine-tunes mBERT on parallel corpora using various cross-lingual objectives, and achieves state-of-the-art performance on word alignment tasks across many language pairs. We also include Levy et al. (2017), Luong et al. (2015a), and Sabet et al. (2020) as recent word embedding baselines, which we denote as **SENTID**, **BIVEC** and **BIS2V**, respectively. All of these baselines are very similar in terms of methodology: SENTID trains a Skip-Gram model that predicts a sentence ID (which is assigned to each set of parallel sentences) from the component words; BIVEC trains a Skip-Gram model that predicts the context cross-lingually based on the word-alignment information;[13] and BIS2V trains a Continuous Bag-of-Words (CBOW) model that predicts a target word from the rest of the sentence and its parallel sentence. Sabet et al. (2020) and Marie and Fujita (2019) show that these joint learning models perform better than mapping-based methods, which align monolingual word embeddings cross-lingually.[14] Regarding the vocabulary size and word embedding dimension, we always use the same values for all the baselines and our model, to ensure fairness.[15]

In addition to these neural baselines, we also compare our model against statistical word alignment methods, namely GIZA++ (Och and Ney, 2003) and Fast Align (Dyer et al., 2013). These are pre-neural methods based on the IBM models (Brown et al., 1993), and still serve as de facto standard models to generate word alignments (Cao et al., 2020; Aldarmaki and Diab, 2019). For all the baselines, we use the authors' implementations.[16]

---

[10] We use white space as the word delimiter and keep all tones in Na sentences, as removing tones increases polysemy in the bilingual dictionary we use for evaluation. For Chinese, we perform word segmentation using the Stanford Word Segmenter (Chang et al., 2008) after data cleaning.

[11] Recently, Bustamante et al. (2020) attempted to scrape monolingual data from PDF documents, but the size of the resulting data is still too small (22k sentences) to apply the latest pretraining methods.

[12] We used the bert-base-multilingual-cased model, following the original paper.

[13] We use Fast Align to generate the alignment.

[14] Besides, Wada et al. (2019) show that the mapping models perform very poorly on low-resource conditions. Based on these findings, we did not include them as our baselines.

[15] Except for the mBERT baseline, which has its pre-defined vocabulary and word embedding dimension, i.e. 768.

### 3.3 Experimental Settings and Evaluation

In our experiments, we train cross-lingual embeddings for five low-resource language pairs: Griko–Italian, Shipibo-Konibo–Spanish and Na–{French, Chinese, English}. For the Griko–Italian pair, we evaluate models on a cross-lingual word alignment task and report alignment accuracy (1−AER). We use the gold alignments manually annotated over the 330 Griko–Italian sentences. To produce alignments using Giza++ and Fast Align, we train them on the 330 sentences with or without additional 10k sentences from a second corpus,[17] and combine forward and backward alignments using the grow-diag-final-and heuristic. For the word embedding-based methods, we train them on the same data, and align each word in a sentence to the closest word in its translation using static or contextualised word embeddings.[18] To calculate word similarity, we use cross-domain similarity local scaling (Conneau et al., 2018):

$$\text{CSLS}(x, y) = 2\cos(x, y) - \frac{1}{K}\sum_{y_t \in \mathcal{N}_T(x)} \cos(x, y_t) \\ - \frac{1}{K}\sum_{x_t \in \mathcal{N}_S(y)} \cos(x_t, y),$$

where $\cos(x, y)$ denotes cosine similarity between $x$ and $y$, and $\mathcal{N}_T(x)$ and $\mathcal{N}_S(y)$ denote the $K$ nearest words to $x$ or $y$ in a target or source sentence; we set $K$ to 3 in the word alignment task. For the mBERT baseline, we follow the authors in using the softmax function.[19]

For the Shipibo-Konibo–Spanish and Na–{French, Chinese, English} pairs, we perform bilingual lexicon induction (BLI). That is, for each source word in a bilingual dictionary, we extract the $k$ nearest words from the whole target vocabulary and see whether they are listed as translations in the dictionary. We set $k$ to 1 or 5, and report P@1 and P@5. For evaluation, we use a Shipibo-Konibo–Spanish dictionary[20] (Maguiño-Valencia et al., 2018) and Na–French–Chinese–English dictionaries (Michaud, 2018). Based on extracting words that are present in the parallel corpora, we identified 79, 262, 215 and 87 word pairs for Shipibo-Konibo–Spanish, Na–French, Na–Chinese, and Na–English.[21] To perform BLI with GIZA++ and Fast Align, we use their source-to-target probability table. We also try using the result of bidirectional word alignments, aligning each word to the most frequently aligned words to it.[22] For the neural baselines and our model, we use static word embeddings and employ CSLS to measure the word embedding similarities. To obtain static word embeddings using mBERT, we calculate the contextualised representations for each word (calculated as the average of its subword embeddings), and take the average over all word occurrences.[23] In BLI, $\mathcal{N}_T(x)$ and $\mathcal{N}_S(y)$ denote the $K$ closest words extracted from the whole vocabulary, with $K = 10$, following Conneau et al. (2018).

### 3.4 Model Selection

Since our evaluation data (i.e. bilingual dictionaries and gold word alignments) is extremely limited, we do not have access to validation data to perform model selection over. Therefore, for all methods except ours, we run the models with different configurations and **report the best scores of the baselines on the test data to show their upper-bound performance**, which clearly gives a significant advantage to the baselines.[24] For Fast Align and GIZA++, we train the models for 5 (default), 10, 15, or 20 iterations independently, and report the best score.[25] For the neural baselines, we evaluate each model-checkpoint and report the best score; we fine-tune the mBERT baseline for 40,000 steps[26] with 20 checkpoints, and train SENTID and BIS2V for 1,000 epochs with 100 checkpoints to ensure convergence. For BIVEC, we increase the training corpus size by 20 times by duplicating the sentences and train the model for 50 epochs with 50 checkpoints.[27]

---

[16]Fast Align: https://github.com/clab/fast_align, Giza++: https://github.com/moses-smt/giza-pp, SENTID: https://bitbucket.org/omerlevy/xling_embeddings/src/default, BIVEC: https://github.com/lmthang/bivec, BIS2V https://github.com/epfml/Bi-Sent2Vec, and mBERT https://github.com/neulab/awesome-align

[17]This is a standard training and evaluation setup for these unsupervised word alignment models: the main objective of word alignment is to extract equivalent words from parallel texts, not to perform translation on held-out data.

[18]We align words bidirectionally and use the grow-diag-final-and heuristic to produce the final result.

[19]We also experimented with the CSLS method and got comparable results.

[20]In this dictionary, sets of synonyms are aligned cross-lingually and we regard each member of them as translations.

[21]Since the Shipibo-Konibo–Spanish parallel corpora contain pairs of words as well as sentences, we include them in the evaluation data and remove them from training data.

[22]We use the probability tables as a backup when there are less than $k$ aligned words.

[23]We also tried taking the average of the static subword embeddings of mBERT, but observed much worse results.

[24]In addition, we tuned the hyper-parameters of BIS2V and BIVEC based on P@1 on the na–en test data, based on the observation that they were very sensitive to hyper-parameters in low-resource conditions (e.g. BIVEC ranged from 5.4 to 33.8 P@1 in the na–en BLI task). Refer to Appendix C for the hyper-parameters of the baselines we used in our experiments.

| src–tgt | P@$k$ | mBERT | SENTID | BIVEC | BIS2V | Fast Align | | GIZA++ | | OURS | | |
|---|---|---|---|---|---|---|---|---|---|---|---|---|
| | | MLM | SG | SG | CBOW | Ptable | +Align | Ptable | +Align | Word | +$SW_{ave}$ | +$SW_{cnn}$ |
| nru–en | 1 | 1.8 | 29.7 | 32.9 | 26.6 | 23.0 | 21.6 | 25.7 | 21.6 | 30.2 | 32.0 | **35.6** |
| | 5 | 4.1 | 50.9 | 50.4 | 44.6 | 48.6 | 50.0 | 43.2 | 47.3 | 50.9 | **56.3** | 51.8 |
| nru–fr | 1 | 0.2 | 22.7 | 23.6 | 19.6 | 23.6 | 20.8 | 22.2 | 15.7 | 27.3 | 29.9 | **32.1** |
| | 5 | 0.8 | 37.7 | 40.1 | 30.2 | 35.2 | 36.1 | 32.9 | 37.5 | 43.7 | **47.8** | 46.0 |
| nru–zh | 1 | 0.5 | 28.7 | 31.9 | 25.1 | 31.6 | 29.9 | 29.4 | 25.7 | 31.6 | 36.7 | **38.5** |
| | 5 | 1.2 | 44.8 | 50.6 | 46.0 | 46.0 | 44.4 | 44.4 | 46.5 | 49.6 | **56.1** | 55.4 |
| shp–es | 1 | 12.0 | 25.1 | 29.5 | 26.2 | 31.1 | 32.8 | 34.4 | 26.2 | 35.0 | 35.0 | **35.5** |
| | 5 | 18.0 | 45.9 | 43.7 | 38.8 | 47.5 | 47.5 | 42.6 | 45.9 | 54.6 | 57.9 | **59.0** |

Table 3: The performance on bilingual lexicon induction (BLI). "+Align" indicates bidirectional alignments are used with the probability table (Ptable) as backup. The scores of the neural models are averaged over three runs.

For our model, on the other hand, we use a simple early-stopping criterion that doesn't require external data. First, we build a pseudo bilingual dictionary from the training data. To retrieve pseudo bilingual word pairs, we compute the Dice Coefficient (Dice, 1945; Smadja et al., 1996) and extract pairs of words that appear $\geq 3$ times in each language and whose Dice Coefficient is $\geq 0.8$ across two languages. We perform model selection based on the BLI performance on this pseudo dictionary.

### 3.5 Results

Table 3 shows the results for BLI.[28] We run the neural baselines and our model three times with different seeds and report the average score since neural models can be unstable with little data. It clearly shows that our model outperforms all the baseline models by a large margin for every language pair. It also shows that utilising shared subword embeddings (+$SW_{ave}$ and +$SW_{cnn}$) further improves our model. Compared to the neural baselines, our model performs better even without subword information, demonstrating its efficiency. The mBERT baseline performs very poorly, likely because of its sub-optimal tokenisation for endangered languages. Table 4 compares our bilingual and multilingual models. The multilingual model is trained jointly on the three parallel corpora, sharing parameters among the four languages. The table shows that the multilingual model achieves better performance overall, especially for Na–English and Na–Chinese, where the number of the aligned sentences is much smaller than for Na–French. This result demonstrates that our model is not only able to embed multiple languages into the same space, but also benefits from extra sentences aligned between additional languages.

| src–tgt | Word | | +$SW_{ave}$ | | +$SW_{cnn}$ | |
|---|---|---|---|---|---|---|
| | bi | multi | bi | mult | bi | multi |
| nru–en | 30.2 | **34.2** | 32.0 | **37.8** | 35.6 | **40.5** |
| nru–fr | 27.3 | **28.7** | 29.9 | 29.8 | **32.1** | 30.7 |
| nru–zh | 31.6 | **36.9** | 36.7 | **40.1** | 38.5 | **40.8** |

Table 4: Our model performance (P@1) on BLI when the model is trained on two and four languages ("bi" vs. "multi"). All scores are averaged over three runs.

Table 5 shows the results for the Griko–Italian word alignment task. It shows that our model performs the best of all the models when they are trained on the 330 sentences only. This is particularly surprising given that encoder-decoder models are usually not effective when trained on small-scale data of magnitude 100s of sentence pairs. The result also shows that our model produces much better static word embeddings than SENTID, BIVEC and BIS2V, demonstrating the importance of considering word order information. When we use the additional 10k sentences, the performance of the baselines drops substantially except for mBERT,[29] likely because of the differences in domains and tokenisation schemes, with the smaller Griko corpus closely following Italian norms. On the other hand, our model achieves good results under both conditions, indicating the robustness of our model to noisy real-world data.

---

[25] For GIZA++, we trained HMM and IBM models 1, 3, and 4 with the same number of iterations, following the default setting. We also tune the number of word classes chosen among {10, 20, 30, 40, 50}.

[26] This follows the original paper, and with those steps the model converged well in our experiments as well.

[27] Without this, it performed very poorly, likely because the code is optimised for reasonably-sized data. Moreover, we trained BIVEC using the word alignments produced by the best Fast Align model, and thus it's doubly optimised.

[28] We present an analysis of retrieved words in Appendix D.

| src–tgt | #Sents | mBERT | SENTID | BIVEC | BIS2V | Fast Align | GIZA++ | OURS | | |
|---|---|---|---|---|---|---|---|---|---|---|
| | | MLM | SG | SG | CBOW | Align | Align | Word (static) | Word $+SW_\text{ave}$ | $+SW_\text{cnn}$ |
| grk–it | 330 | 91.1 | 66.7 | 59.2 | 65.3 | 90.2 | 93.3 | 88.0 | 93.1 | 93.4 | **93.5** |
| | +10k | **93.1** | 61.8 | 56.1 | 58.1 | 87.1 | 81.0 | 86.6 | 92.6 | 93.0 | 93.0 |

Table 5: The performance (1−AER) on the Griko-Italian word alignment task with or without additional 10k parallel sentences. "Word (static)" is the result when our model uses static word embeddings instead of contextualised ones. The scores of the neural models are averaged over three runs.

### 3.6 Results on High-Resource Languages

To investigate how our model performs on high-resource conditions, we conduct additional word-alignment experiments on four high-resource language pairs: Japanese-English (ja-en), English-Inuktitut (en-iu), German-English (de-en), and English-French (en-fr). Regarding Inuktitut, there is no large-scale monolingual data, making it a salient test case for our model. We use benchmark word-alignment data sets for each language pair,[30] where the de-en and en-fr data sets contain about 2M and 1M parallel sentences, and the ja-en and en-iu ones about 0.3M. We apply SentencePiece to each corpus[31] and use them to train all the models except for mBERT, for which we use its pre-trained tokeniser. To perform word alignment, first we align subwords and align words if any of their subwords is aligned.[32] We use the same model selection criteria (Section 3.4) to report the upper bound of the baselines.[33]

[29]We conjecture this is because the additional data helped mBERT (esp. its positional embeddings) to learn that these two languages have very similar syntax.
[30]Kyoto Free Translation Task (Neubig, 2011) (ja-en); the Legislative Assembly of Nunavut (Martin et al., 2003) (en-iu); Europarl (Koehn, 2005) (de-en); and Canadian Hansards (Germann, 2001; Mihalcea and Pedersen, 2003) (en-fr).
[31]For the en-iu corpus, we segmented the Inuktitut sentences only, as there is a significant gap between the English and Inuktitut vocabulary size, i.e. 22k vs. 400k.
[32]GIZA++ and Fast Align also benefit from this method.
[33]We train the word embedding baselines for 100 epochs and the mBERT baseline for 40,000 steps with 20 checkpoints, and Fast Align and GIZA++ for 5, 10, 15 or 20 epochs, using 50 word classes (Moses default) for GIZA++.

| Model | en-iu | | | ja-en | | |
|---|---|---|---|---|---|---|
| | P | R | 1−A | P | R | 1−A |
| mBERT | 69.7 | 86.7 | 74.0 | 73.9 | 56.4 | 64.0 |
| SENTID | 72.9 | 86.3 | 75.3 | 57.7 | 46.8 | 51.7 |
| BIVEC | 67.4 | 84.4 | 70.3 | 56.3 | 46.1 | 50.7 |
| BIS2V | 65.2 | 82.8 | 68.4 | 52.3 | 41.3 | 46.2 |
| Fast Align | 80.5 | 96.0 | 82.8 | 61.7 | 51.7 | 56.3 |
| GIZA++ | 85.7 | 93.5 | 87.1 | 71.3 | 55.9 | 62.7 |
| OURS (static) | 73.1 | 88.6 | 75.6 | 59.4 | 48.0 | 53.1 |
| OURS | 83.5 | **99.3** | 85.6 | 62.7 | 65.5 | 64.1 |
| OURS+$SW_\text{ave}$ | 83.0 | 98.8 | 85.2 | 63.0 | **66.9** | 64.9 |
| OURS+$SW_\text{ave}$+null | **91.8** | 97.0 | **92.8** | **75.2** | 59.7 | **66.6** |

Table 6: Precision ("P"), Recall ("R") and 1−AER ("1−A") of word alignment on high-resource conditions. P and R are calculated based on possible and sure alignments, resp. "+null" denotes the result with null alignments.

Table 6 shows the results of the ja-en and en-iu word alignment experiments. It demonstrates that our model (OURS) significantly outperforms the other static word-embedding baselines. It also outperforms mBERT in en-iu and even in ja-en, which is very surprising given that mBERT is pre-trained on large-scale monolingual data for Japanese and English.[34] We also tried training our subword-aware model (OURS+SW$_\text{ave}$) by segmenting subwords into smaller word pieces and learning "subsubword" embeddings. The result shows that this approach improves our model for ja-en but not en-iu, probably because some Japanese characters (e.g. kanji) contain much semantic information. When compared to the word alignment tools, our model outperforms Fast Align and is comparable to GIZA++, achieving lower precision and higher recall. This is because, unlike GIZA++, our simple alignment algorithm based on CSLS cannot handle NULL alignments (untranslatable words) and generates more alignments than necessary. To handle those words, we apply the following heuristic: discard alignments between $x$ and $y$ if $CSLS(x, y) \leq 0$ or $\cos(x, y) \leq \min(\cos(x, BOS), \cos(BOS, y))$. This improves our model substantially ("**+null**" in Table 6) and it outperforms all the baselines.[35]

[34]Inuktitut is not included in the training data of mBERT, accounting for its poor performance on en-iu.

| Model | Method | Training Data | de-en | en-fr |
|---|---|---|---|---|
| Jalili Sabet et al. (2020) | Masked LM | monolingual | 18.8 | 7.6 |
| Fast Align | Statistical Model | bilingual | 27.0 | 10.5 |
| eflomal | Statistical Model | bilingual | 22.6 | 8.2 |
| GIZA++ | Statistical Model | bilingual | 20.6 | 5.9 |
| Zenkel et al. (2020) | Pre-trained NMT+Aligner | bilingual | 16.0 | 5.0 |
| Chen et al. (2020) | Pre-trained NMT+Aligner | bilingual | 15.4 | 4.7 |
| Dou and Neubig (2021)+α-entmax | Fine-tuned Masked LM | monolingual+bilingual | 16.1 | **4.1** |
| Dou and Neubig (2021)+sotmax | Fine-tuned Masked LM | monolingual+bilingual | 15.6 | 4.4 |
| Dou and Neubig (2021)+α-entmax (β = 1) | Fine-tuned Masked LM | monolingual+multilingual | 15.0 | 4.5 |
| Dou and Neubig (2021)+sotmax (β = 1) | Fine-tuned Masked LM | monolingual+multilingual | 15.1 | 4.5 |
| OURS | NMT | bilingual | 16.4 | 9.1 |
| OURS +null | NMT | bilingual | **14.0** | 4.5 |

Table 7: Comparison of AER scores among various word alignment models. All the scores except for ours are cited from Dou and Neubig (2021). "+null" denotes the result with null alignments.

Lastly, Table 7 shows the results of the de-en and en-fr experiments. We cite the scores of the baselines from Dou and Neubig (2021), and report AER instead of 1−AER following the original table. It shows that "OURS +null" performs comparably to mBERT for en-fr (4.5 vs. 4.1), and outperforms it for de-en (14.0 vs. 15.0), establishing a new state-of-the-art with much less data and fewer parameters. The table also shows that our method is much simpler than the other NMT-based models (Zenkel et al., 2020; Chen et al., 2020), which pre-train an NMT model and then train an alignment model on top of it. Another important difference is that our model can produce cross-lingual representations while the NMT-based baselines can generate word alignments only.

### 3.7 Alignment with Pre-trained Embeddings

To employ large-scale monolingual data, we try initialising word embeddings of a high-resource language with pre-trained word embeddings, and train the word embeddings of a low-resource language in the same embedding space. During training, we freeze the pre-trained embeddings $E_{pre}^{\ell}$ and apply the element-wise operation $a \otimes E_{pre}^{\ell} + b$, where $a$ and $b$ are trainable vectors and shared among all the words in $E_{pre}^{\ell}$. For the other words, we train subword-aware embeddings ($SW_{ave}$) from scratch.

---
[35]For grk-it, the performance (1−AER) slightly dropped, e.g. "+$SW_{cnn}$+null" achieved 92.3/93.2 w/w.o the 10k sentences, likely because there are very few NULL alignments.

| Model | Source | Retrieved Words (Top 3) |
|---|---|---|
| BIVEC | ʐo˩ (lunch, noon) | **lunch**, woman, lunchtime |
| OURS | | **lunch**, lunchtime, outside |
| + Pre | | **lunch**, lunchtime, dinner |
| BIVEC | tʂʰɯ˧-qo˧ (here) | **here**, fireplace, beam |
| OURS | | **here**, sat, there |
| + Pre | | **here**, there, where |
| BIVEC | kwɑ˧tʂʰɑ˧ (coffin) | **coffin**, rush, near |
| OURS | | **coffin**, rush, placed |
| + Pre | | **coffin**, cremated, corpse |

Table 8: Examples of retrieved words on nru-en BLI. "+ Pre" denotes the use of pre-trained embeddings.

We conduct an experiment on nru-en, where we pre-train English word embeddings using FastText on 10M sentences sampled from web-crawled data, OSCAR (Ortiz Suárez et al., 2020). The model achieves 30.6/52.3 on P@1/5, underperforming OURS+$SW_{ave}$ without pre-training (32.0/56.3), possibly because of the domain difference between the parallel and monolingual data. However, a closer look at the matched words reveals that pre-training can improve the retrieval performance in several cases, as shown in Table 8 (more examples are in Appendix D). It shows that even though all the models successfully match the correct words, our models retrieve more relevant words to the target word, especially when trained with pre-trained embeddings. This suggests that pre-training may benefit the model on other semantic tasks. Pre-training also makes it possible to measure the similarities between Na words and English words that are out-of-vocabulary in the parallel corpus.

| Model | nru | | | | ja-en |
|---|---|---|---|---|---|
| | en | fr | zh | multi | |
| OURS+$SW_{ave}$ (+null) | **32.0** | **29.9** | 36.7 | **35.9** | **66.6** |
| −(sub)subword | 30.2 | 27.3 | 31.6 | 33.3 | 65.9 |
| −$E^{\ell}_{w_i}$ in Eqn. (8) | 27.5 | 21.4 | **40.8** | 28.2 | 59.7 |
| −Bkw ($\overleftarrow{g}$) | 30.2 | 28.6 | 34.9 | 33.1 | 64.8 |
| −dropout | 27.9 | 25.1 | 29.6 | 27.0 | 60.3 |
| −weight tying | 24.3 | 23.4 | 29.0 | 32.5 | 63.9 |
| −reconstuction | 0.9 | 1.2 | 0.4 | 34.8 | 36.5 |

Table 9: Ablation results for our model (P@1 (nru) or 1−AER (ja-en)). "multi" indicates the average P@1 of our multilingual model over the three language pairs.

### 3.8 Ablation Studies

To investigate the effectiveness of our model, we perform ablation studies, targeting: shared subword embeddings, word-specific embeddings $E^{\ell}_{w_i}$ in Eqn. (8), backward decoding, dropout, weight tying, and the reconstruction objective. Table 9 shows the results. Without the reconstruction loss (i.e. the standard NMT model), the model performs very poorly in the bilingual settings. In the multilingual setting, however, the reconstruction is not essential because the model is trained to translate multiple languages into Na, which forces the source languages to be encoded in the same space. The table also shows that weight tying is very effective. We find this result intriguing, as usually it does not affect translation quality very much (Press and Wolf, 2017). In our model, however, weight tying is crucial in two ways: first, it reduces the number of parameters and prevents the model from learning word embeddings in different spaces; and second, it introduces more direct connections between source and target (sub)word embeddings through attention. Aggressive dropout is also effective in all the conditions, preventing the model from learning language-specific embeddings using different spaces. Word-specific embeddings also improve the model performance except for nru-zh.[36] Lastly, backward decoding also improves performance, incorporating right-to-left contexts into the word embeddings.

### 4 Related Work

There are two main approaches to learning cross-lingual word embeddings. One is to learn a matrix that aligns pretrained monolingual embeddings.

---

[36]This is likely because Chinese words can be represented well by their component characters (Li et al., 2019). However, in "mutli", $E^{\ell}_{w_i}$ is beneficial for nru-zh as well.

Most such methods exploit bilingual dictionaries to learn the mapping matrix (Mikolov et al., 2013b; Xing et al., 2015; Joulin et al., 2018), but recently a number of methods have succeeded in learning the matrix without supervision (Zhang et al., 2017; Conneau et al., 2018; Artetxe et al., 2018a). However, Wada et al. (2019) show that this approach does not work well on low-resource conditions. The second approach is to jointly train cross-lingual embeddings in a common space. Most existing methods extend bag-of-words models (e.g. Skip-Gram) to incorporate cross-lingual information provided by parallel or comparable corpora (Hermann and Blunsom, 2014; Vulic and Moens, 2016; Levy et al., 2017; Dufter et al., 2018a,b; Luong et al., 2015a; Sabet et al., 2020; Sarioglu Kayi et al., 2020), or bilingual dictionaries (Gouws and Søgaard, 2015; Duong et al., 2016). Recently, masked language models such as XLM (Conneau and Lample, 2019) and mBERT (Devlin et al., 2019) have been shown to generate cross-lingual representations without parallel data, but require an enormous amount of monolingual data, which is not available for endangered languages.

Similar to our work, some papers use multilingual neural machine translation models to obtain cross-lingual representations (Eriguchi et al., 2018; Schwenk and Douze, 2017; Artetxe and Schwenk, 2019; Schwenk, 2018). However, they employ extremely large and/or multilingual data aligned among more than two languages (e.g. Europarl, United Nations). Another important difference is that their methods focus on learning cross-lingual *sentence* representations only (i.e. not at the word level), and are evaluated on cross-lingual sentence retrieval or sentence classification tasks.

### 5 Conclusion

We propose a new approach for learning contextualised cross-lingual word embeddings that can be trained with a tiny parallel corpus. We evaluate models on real-world data for three endangered languages, and also on benchmark data sets for four high-resource languages, and show that our model outperforms existing methods at bilingual lexicon induction and word alignment.

### 6 Acknowledgement

We are grateful to Oliver Adams and Alexis Michaud for providing us with the pre-processing code and detailed instructions for Na data.

# References

Oliver Adams, Adam Makarucha, Graham Neubig, Steven Bird, and Trevor Cohn. 2017. Cross-lingual word embeddings for low-resource language modeling. In *Proceedings of the 15th Conference of the European Chapter of the Association for Computational Linguistics: Volume 1, Long Papers*, pages 937–947, Valencia, Spain. Association for Computational Linguistics.

Hanan Aldarmaki and Mona Diab. 2019. Context-aware cross-lingual mapping. In *Proceedings of the 2019 Conference of the North American Chapter of the Association for Computational Linguistics: Human Language Technologies, Volume 1 (Long and Short Papers)*, pages 3906–3911, Minneapolis, Minnesota. Association for Computational Linguistics.

Antonios Anastasopoulos, Marika Lekakou, Josep Quer, Eleni Zimianiti, Justin DeBenedetto, and David Chiang. 2018. Part-of-speech tagging on an endangered language: a parallel Griko-Italian resource. In *Proceedings of the 27th International Conference on Computational Linguistics*, pages 2529–2539, Santa Fe, New Mexico, USA. Association for Computational Linguistics.

Mikel Artetxe, Gorka Labaka, and Eneko Agirre. 2018a. A robust self-learning method for fully unsupervised cross-lingual mappings of word embeddings. In *Proceedings of the 56th Annual Meeting of the Association for Computational Linguistics (Volume 1: Long Papers)*, pages 789–798, Melbourne, Australia. Association for Computational Linguistics.

Mikel Artetxe, Gorka Labaka, Eneko Agirre, and Kyunghyun Cho. 2018b. Unsupervised neural machine translation. In *Proceedings of the 6th International Conference on Learning Representations*, Vancouver, BC, Canada.

Mikel Artetxe and Holger Schwenk. 2019. Massively multilingual sentence embeddings for zero-shot cross-lingual transfer and beyond. *Transactions of the Association for Computational Linguistics*, 7:597–610.

Zhenisbek Assylbekov and Rustem Takhanov. 2018. Reusing weights in subword-aware neural language models. In *Proceedings of the 2018 Conference of the North American Chapter of the Association for Computational Linguistics: Human Language Technologies, Volume 1 (Long Papers)*, pages 1413–1423, New Orleans, Louisiana. Association for Computational Linguistics.

Piotr Bojanowski, Edouard Grave, Armand Joulin, and Tomas Mikolov. 2017. Enriching word vectors with subword information. *Transactions of the Association for Computational Linguistics*, 5:135–146.

Peter F. Brown, Stephen A. Della Pietra, Vincent J. Della Pietra, and Robert L. Mercer. 1993. The mathematics of statistical machine translation: Parameter estimation. *Computational Linguistics*, 19(2):263–311.

Gina Bustamante, Arturo Oncevay, and Roberto Zariquiey. 2020. No data to crawl? monolingual corpus creation from PDF files of truly low-resource languages in Peru. In *Proceedings of the 12th Language Resources and Evaluation Conference*, pages 2914–2923, Marseille, France. European Language Resources Association.

Steven Cao, Nikita Kitaev, and Dan Klein. 2020. Multilingual alignment of contextual word representations. In *Proceedings of the 8th International Conference on Learning Representations*, Virtual Conference, Formerly Addis Ababa, Ethiopia.

Pi-Chuan Chang, Michel Galley, and Christopher D. Manning. 2008. Optimizing Chinese word segmentation for machine translation performance. In *Proceedings of the Third Workshop on Statistical Machine Translation*, pages 224–232, Columbus, Ohio. Association for Computational Linguistics.

Yun Chen, Yang Liu, Guanhua Chen, Xin Jiang, and Qun Liu. 2020. Accurate word alignment induction from neural machine translation. In *Proceedings of the 2020 Conference on Empirical Methods in Natural Language Processing (EMNLP)*, pages 566–576, Online. Association for Computational Linguistics.

Alexis Conneau and Guillaume Lample. 2019. Cross-lingual language model pretraining. In H. Wallach, H. Larochelle, A. Beygelzimer, F. d'Alché-Buc, E. Fox, and R. Garnett, editors, *Advances in Neural Information Processing Systems 32*, pages 7059–7069. Curran Associates, Inc.

Alexis Conneau, Guillaume Lample, Marc'Aurelio Ranzato, Ludovic Denoyer, and Hervé Jégou. 2018. Word translation without parallel data. In *Proceedings of the 6th International Conference on Learning Representations*, Vancouver, BC, Canada.

Jacob Devlin, Ming-Wei Chang, Kenton Lee, and Kristina Toutanova. 2019. BERT: Pre-training of deep bidirectional transformers for language understanding. In *Proceedings of the 2019 Conference of the North American Chapter of the Association for Computational Linguistics: Human Language Technologies, Volume 1 (Long and Short Papers)*, pages 4171–4186, Minneapolis, Minnesota. Association for Computational Linguistics.

Lee R. Dice. 1945. Measures of the amount of ecologic association between species. *Ecology*, 26(3):297–302.

Thi-Ngoc-Diep Do, Alexis Michaud, and Eric Castelli. 2014. Towards the automatic processing of Yongning Na (Sino-Tibetan): developing a 'light' acoustic model of the target language and testing 'heavyweight' models from five national languages. In *4th International Workshop on Spoken Language Technologies for Under-resourced Languages (SLTU 2014)*, pages 153–160, St Petersburg, Russia.

Zi-Yi Dou and Graham Neubig. 2021. Word alignment by fine-tuning embeddings on parallel corpora. In *Proceedings of the 16th Conference of the European Chapter of the Association for Computational Linguistics: Main Volume*, pages 2112–2128, Online. Association for Computational Linguistics.

Philipp Dufter, Mengjie Zhao, Martin Schmitt, Alexander Fraser, and Hinrich Schütze. 2018a. Embedding learning through multilingual concept induction. In *Proceedings of the 56th Annual Meeting of the Association for Computational Linguistics (Volume 1: Long Papers)*, pages 1520–1530, Melbourne, Australia. Association for Computational Linguistics.

Philipp Dufter, Mengjie Zhao, and Hinrich Schütze. 2018b. Multilingual embeddings jointly induced from contexts and concepts: Simple, strong and scalable. *CoRR*, abs/1811.00586.

Long Duong, Hiroshi Kanayama, Tengfei Ma, Steven Bird, and Trevor Cohn. 2016. Learning crosslingual word embeddings without bilingual corpora. In *Proceedings of the 2016 Conference on Empirical Methods in Natural Language Processing*, pages 1285–1295, Austin, Texas. Association for Computational Linguistics.

Chris Dyer, Victor Chahuneau, and Noah A. Smith. 2013. A simple, fast, and effective reparameterization of IBM model 2. In *Proceedings of the 2013 Conference of the North American Chapter of the Association for Computational Linguistics: Human Language Technologies*, pages 644–648, Atlanta, Georgia. Association for Computational Linguistics.

Akiko Eriguchi, Melvin Johnson, Orhan Firat, Hideto Kazawa, and Wolfgang Macherey. 2018. Zero-shot cross-lingual classification using multilingual neural machine translation. *CoRR*, abs/1809.04686.

Ana-Paula Galarreta, Andrés Melgar, and Arturo Oncevay. 2017. Corpus creation and initial SMT experiments between Spanish and Shipibo-Konibo. In *Proceedings of the International Conference Recent Advances in Natural Language Processing, RANLP 2017*, pages 238–244, Varna, Bulgaria. INCOMA Ltd.

U. Germann. 2001. Aligned hansards of the 36th parliament of canada.

Stephan Gouws and Anders Søgaard. 2015. Simple task-specific bilingual word embeddings. In *Proceedings of the 2015 Conference of the North American Chapter of the Association for Computational Linguistics: Human Language Technologies*, pages 1386–1390, Denver, Colorado. Association for Computational Linguistics.

Karl Moritz Hermann and Phil Blunsom. 2014. Multilingual models for compositional distributed semantics. In *Proceedings of the 52nd Annual Meeting of the Association for Computational Linguistics (Volume 1: Long Papers)*, pages 58–68, Baltimore, Maryland. Association for Computational Linguistics.

Sepp Hochreiter and Jürgen Schmidhuber. 1997. Long short-term memory. *Neural Computation*, 9(8):1735–1780.

Junjie Hu, Sebastian Ruder, Aditya Siddhant, Graham Neubig, Orhan Firat, and Melvin Johnson. 2020. Xtreme: A massively multilingual multi-task benchmark for evaluating cross-lingual generalization. *CoRR*, abs/2003.11080.

Hakan Inan, Khashayar Khosravi, and Richard Socher. 2017. Tying word vectors and word classifiers: A loss framework for language modeling. In *Proceedings of the 5th International Conference on Learning Representations*, Toulon, France.

Masoud Jalili Sabet, Philipp Dufter, François Yvon, and Hinrich Schütze. 2020. SimAlign: High quality word alignments without parallel training data using static and contextualized embeddings. In *Findings of the Association for Computational Linguistics: EMNLP 2020*, pages 1627–1643, Online. Association for Computational Linguistics.

Armand Joulin, Piotr Bojanowski, Tomas Mikolov, Hervé Jégou, and Edouard Grave. 2018. Loss in translation: Learning bilingual word mapping with a retrieval criterion. In *Proceedings of the 2018 Conference on Empirical Methods in Natural Language Processing*, pages 2979–2984, Brussels, Belgium. Association for Computational Linguistics.

Diederik P. Kingma and Jimmy Ba. 2015. Adam: A method for stochastic optimization. In *Proceedings of the 3rd International Conference on Learning Representations*.

Philipp Koehn. 2005. Europarl: A Parallel Corpus for Statistical Machine Translation. In *Conference Proceedings: the tenth Machine Translation Summit*, Phuket, Thailand. Asia-Pacific Association for Machine Translation.

Taku Kudo and John Richardson. 2018. SentencePiece: A simple and language independent subword tokenizer and detokenizer for neural text processing. In *Proceedings of the 2018 Conference on Empirical Methods in Natural Language Processing: System Demonstrations*, pages 66–71, Brussels, Belgium. Association for Computational Linguistics.

Guillaume Lample, Alexis Conneau, Ludovic Denoyer, and Marc'Aurelio Ranzato. 2018. Unsupervised machine translation using monolingual corpora only. In *Proceedings of the 6th International Conference on Learning Representations*, Vancouver, BC, Canada.

Omer Levy, Anders Søgaard, and Yoav Goldberg. 2017. A strong baseline for learning cross-lingual word embeddings from sentence alignments. In *Proceedings of the 15th Conference of the European Chapter of the Association for Computational Linguistics: Volume 1, Long Papers*, pages 765–774, Valencia, Spain. Association for Computational Linguistics.

Xiaoya Li, Yuxian Meng, Xiaofei Sun, Qinghong Han, Arianna Yuan, and Jiwei Li. 2019. Is word segmentation necessary for deep learning of Chinese representations? In *Proceedings of the 57th Annual Meeting of the Association for Computational Linguistics*, pages 3242–3252, Florence, Italy. Association for Computational Linguistics.

Thang Luong, Hieu Pham, and Christopher D. Manning. 2015a. Bilingual word representations with monolingual quality in mind. In *Proceedings of the 1st Workshop on Vector Space Modeling for Natural Language Processing*, pages 151–159, Denver, Colorado. Association for Computational Linguistics.

Thang Luong, Hieu Pham, and Christopher D. Manning. 2015b. Effective approaches to attention-based neural machine translation. In *Proceedings of the 2015 Conference on Empirical Methods in Natural Language Processing*, pages 1412–1421, Lisbon, Portugal. Association for Computational Linguistics.

Diego Maguiño-Valencia, Arturo Oncevay-Marcos, and Marco A. Sobrevilla Cabezudo. 2018. WordNet-shp: Towards the building of a lexical database for a Peruvian minority language. In *Proceedings of the Eleventh International Conference on Language Resources and Evaluation (LREC-2018)*, Miyazaki, Japan. European Languages Resources Association (ELRA).

Benjamin Marie and Atsushi Fujita. 2019. Unsupervised joint training of bilingual word embeddings. In *Proceedings of the 57th Annual Meeting of the Association for Computational Linguistics*, pages 3224–3230, Florence, Italy. Association for Computational Linguistics.

Joel Martin, Howard Johnson, Benoit Farley, and Anna Maclachlan. 2003. Aligning and using an English-Inuktitut parallel corpus. In *Proceedings of the HLT-NAACL 2003 Workshop on Building and Using Parallel Texts: Data Driven Machine Translation and Beyond*, pages 115–118.

Boyd Michailovsky, Martine Mazaudon, Alexis Michaud, Séverine Guillaume, Alexandre François, and Evangelia Adamou. 2014. Documenting and researching endangered languages: the Pangloss Collection. In *Language Documentation & Conservation*, volume 8, pages 119–135. "University of Hawaii Press".

Alexis Michaud. 2017. *Tone in Yongning Na: lexical tones and morphotonology*, volume 13 of *Studies in Diversity Linguistics*. Language Science Press.

Alexis Michaud. 2018. Na (Mosuo)-English-Chinese dictionary. https://halshs.archives-ouvertes.fr/halshs-01204638, Version 1.2.

Rada Mihalcea and Ted Pedersen. 2003. An evaluation exercise for word alignment. In *Proceedings of the HLT-NAACL 2003 Workshop on Building and Using Parallel Texts: Data Driven Machine Translation and Beyond*, pages 1–10.

Tomas Mikolov, Kai Chen, Greg Corrado, and Jeffrey Dean. 2013a. Efficient estimation of word representations in vector space. In *Proceedings of the 1st International Conference on Learning Representations (Workshop)*, Scottsdale, Arizona, USA.

Tomas Mikolov, Quoc V. Le, and Ilya Sutskever. 2013b. Exploiting similarities among languages for machine translation. *CoRR*, abs/1309.4168.

Christopher Moseley, editor. 2010. *Atlas of the World's Languages in Danger, 3rd edn.* UNESCO Publishing, Paris. Online version: http://www.unesco.org/culture/en/endangeredlanguages/atlas.

Graham Neubig. 2011. The Kyoto free translation task. http://www.phontron.com/kftt.

Toan Nguyen and David Chiang. 2018. Improving lexical choice in neural machine translation. In *Proceedings of the 2018 Conference of the North American Chapter of the Association for Computational Linguistics: Human Language Technologies, Volume 1 (Long Papers)*, pages 334–343, New Orleans, Louisiana. Association for Computational Linguistics.

Franz Josef Och and Hermann Ney. 2003. A systematic comparison of various statistical alignment models. *Computational Linguistics*, 29(1):19–51.

Pedro Javier Ortiz Suárez, Laurent Romary, and Benoît Sagot. 2020. A monolingual approach to contextualized word embeddings for mid-resource languages. In *Proceedings of the 58th Annual Meeting of the Association for Computational Linguistics*, pages 1703–1714, Online. Association for Computational Linguistics.

Adam Paszke, Sam Gross, Francisco Massa, Adam Lerer, James Bradbury, Gregory Chanan, Trevor Killeen, Zeming Lin, Natalia Gimelshein, Luca Antiga, Alban Desmaison, Andreas Kopf, Edward Yang, Zachary DeVito, Martin Raison, Alykhan Tejani, Sasank Chilamkurthy, Benoit Steiner, Lu Fang, Junjie Bai, and Soumith Chintala. 2019. Pytorch: An imperative style, high-performance deep learning library. In H. Wallach, H. Larochelle, A. Beygelzimer, F. d'Alché-Buc, E. Fox, and R. Garnett, editors, *Advances in Neural Information Processing Systems 32*, pages 8024–8035. Curran Associates, Inc.

Matthew Peters, Mark Neumann, Mohit Iyyer, Matt Gardner, Christopher Clark, Kenton Lee, and Luke Zettlemoyer. 2018. Deep contextualized word representations. In *Proceedings of the 2018 Conference of the North American Chapter of the Association for Computational Linguistics: Human Language Technologies, Volume 1 (Long Papers)*, pages 2227–2237,

New Orleans, Louisiana. Association for Computational Linguistics.

Ofir Press and Lior Wolf. 2017. Using the output embedding to improve language models. In *Proceedings of the 15th Conference of the European Chapter of the Association for Computational Linguistics: Volume 2, Short Papers*, pages 157–163, Valencia, Spain. Association for Computational Linguistics.

Ali Sabet, Prakhar Gupta, Jean-Baptiste Cordonnier, Robert West, and Martin Jaggi. 2020. Robust cross-lingual embeddings from parallel sentences. *ArXiv 1912.12481*.

Efsun Sarioglu Kayi, Vishal Anand, and Smaranda Muresan. 2020. MultiSeg: Parallel data and subword information for learning bilingual embeddings in low resource scenarios. In *Proceedings of the 1st Joint Workshop on Spoken Language Technologies for Under-resourced languages (SLTU) and Collaboration and Computing for Under-Resourced Languages (CCURL)*, pages 97–105, Marseille, France. European Language Resources association.

Holger Schwenk. 2018. Filtering and mining parallel data in a joint multilingual space. In *Proceedings of the 56th Annual Meeting of the Association for Computational Linguistics (Volume 2: Short Papers)*, pages 228–234, Melbourne, Australia. Association for Computational Linguistics.

Holger Schwenk and Matthijs Douze. 2017. Learning joint multilingual sentence representations with neural machine translation. In *Proceedings of the 2nd Workshop on Representation Learning for NLP*, pages 157–167, Vancouver, Canada. Association for Computational Linguistics.

Frank Smadja, Kathleen R. McKeown, and Vasileios Hatzivassiloglou. 1996. Translating collocations for bilingual lexicons: A statistical approach. *Computational Linguistics*, 22(1):1–38.

Nitish Srivastava, Geoffrey Hinton, Alex Krizhevsky, Ilya Sutskever, and Ruslan Salakhutdinov. 2014. Dropout: A simple way to prevent neural networks from overfitting. *Journal of Machine Learning Research*, 15:1929–1958.

Pilar M. Valenzuela. 1997. Basic verb types and argument structures in Shipibo-Conibo. Master's thesis, University of Oregon.

Alonso Vasquez, Renzo Ego Aguirre, Candy Angulo, John Miller, Claudia Villanueva, Željko Agić, Roberto Zariquiey, and Arturo Oncevay. 2018. Toward Universal Dependencies for Shipibo-Konibo. In *Proceedings of the Second Workshop on Universal Dependencies (UDW 2018)*, pages 151–161, Brussels, Belgium. Association for Computational Linguistics.

Ashish Vaswani, Noam Shazeer, Niki Parmar, Jakob Uszkoreit, Llion Jones, Aidan N Gomez, Ł ukasz Kaiser, and Illia Polosukhin. 2017. Attention is all you need. In I. Guyon, U. V. Luxburg, S. Bengio, H. Wallach, R. Fergus, S. Vishwanathan, and R. Garnett, editors, *Advances in Neural Information Processing Systems 30*, pages 5998–6008. Curran Associates, Inc.

Ivan Vulic and Marie-Francine Moens. 2016. Bilingual distributed word representations from document-aligned comparable data. *Journal of Artificial Intelligence Research*, 55(1):953–994.

Takashi Wada, Tomoharu Iwata, and Yuji Matsumoto. 2019. Unsupervised multilingual word embedding with limited resources using neural language models. In *Proceedings of the 57th Annual Meeting of the Association for Computational Linguistics*, pages 3113–3124, Florence, Italy. Association for Computational Linguistics.

Jiateng Xie, Zhilin Yang, Graham Neubig, Noah A. Smith, and Jaime Carbonell. 2018. Neural cross-lingual named entity recognition with minimal resources. In *Proceedings of the 2018 Conference on Empirical Methods in Natural Language Processing*, pages 369–379, Brussels, Belgium. Association for Computational Linguistics.

Chao Xing, Dong Wang, Chao Liu, and Yiye Lin. 2015. Normalized word embedding and orthogonal transform for bilingual word translation. In *Proceedings of the 2015 Conference of the North American Chapter of the Association for Computational Linguistics: Human Language Technologies*, pages 1006–1011. Association for Computational Linguistics.

Marcely Zanon Boito, Antonios Anastasopoulos, Aline Villavicencio, Laurent Besacier, and Marika Lekakou. 2018. A Small Griko-Italian Speech Translation Corpus. In *Proceedings of the 6th International Workshop on Spoken Language Technologies for Under-Resourced Languages*, pages 29–31, Gurugram, India. International Speech Communication Association.

Thomas Zenkel, Joern Wuebker, and John DeNero. 2020. End-to-end neural word alignment outperforms GIZA++. In *Proceedings of the 58th Annual Meeting of the Association for Computational Linguistics*, pages 1605–1617, Online. Association for Computational Linguistics.

Meng Zhang, Yang Liu, Huanbo Luan, and Maosong Sun. 2017. Adversarial training for unsupervised bilingual lexicon induction. In *Proceedings of the 55th Annual Meeting of the Association for Computational Linguistics (Volume 1: Long Papers)*, pages 1959–1970, Vancouver, Canada. Association for Computational Linguistics.

Shiyue Zhang, Benjamin Frey, and Mohit Bansal. 2020. ChrEn: Cherokee-English machine translation for endangered language revitalization. In *Proceedings of the 2020 Conference on Empirical Methods in Natural Language Processing (EMNLP)*, pages 577–595, Online. Association for Computational Linguistics.

| Parameter | Resource Condition | | |
|---|---|---|---|
| | low | medium | high |
| Epoch | 200/100 | 10 | 10 |
| Batch Size | 16 | 80 | 80 |
| Enc LSTM Layers | 1 | 2 | 3 |
| Dec LSTM Layers | 1 | 1 | 2 |
| Dropout Rate | 0.5 | 0.5 | 0.5 |
| Gradient Clipping | 5 | 5 | 5 |
| LSTM Hidden State | 500 | 500 | 768 |
| Word Embedding | 500 | 500 | 768 |
| Subword Embedding | 500 | 500 | – |
| CNN Hidden State | 500 | – | – |
| CNN Window Size | 3 | – | – |

Table 10: Hyper-parameters of our model in low- and high-resource experiments. The first column ("low") denotes the hyper-paramertes for low-resource languages, the second ("medium") for ja-en and iu-en, the last ("high") for de-en and en-fr. CNN is used in low-resource experiments only due to its high computational cost.

## A Details of Our Model

We report the details of our model to ensure reproducibility. Our model was trained using PyTorch (Paszke et al., 2019) on a single GPU.

### A.1 Hyper Parameters

Table 10 shows the hyper-parameters of our model. We tuned the hyper-parameters on low and medium conditions using a subset of the de-en or en-fr data sets. We used the same embedding size in both low and medium resource conditions for simplicity. For very high-resource languages (i.e. de-en and en-fr), we simply increased the number of the encoder and decoder layers by one and set the embedding size to the same as that of mBERT to have a fair comparison. We use Adam (Kingma and Ba, 2015) as the optimiser with the default learning rate. In low-resource experiments, we train our model for 200 epochs in the Na and Griko bilingual experiments, and for 100 epochs for other languages. To learn our subword-aware models, we applied SentencePiece separately to each language with the vocabulary size set to 1,000 (in low-resource experiments) or 1,000 + the number of character types (for ja-en and en-iu).

### A.2 Number of Parameters

Table 11 shows the number of parameters of our model in millions. In low-resource experi-ments, most of the model parameters are located in LSTMs (9.5 millions) because the vocabulary sizes are very small. While our model itself is clearly more complex than the neural baselines (except for mBERT), the dimension of the word embeddings, which our model is trained for, is set to the same for all the models to ensure fairness.

| src-tgt | OURS | | |
|---|---|---|---|
| | Word | $+SW_{\text{ave}}$ | $+SW_{\text{cnn}}$ |
| nru–en | 11.2 | 11.7 | 12.5 |
| nru–zh | 13.3 | 14.2 | 15.0 |
| nru–fr | 16.6 | 17.3 | 18.0 |
| grk–it (s) | 10.3 | 10.8 | 11.5 |
| grk–it (s+l) | 21.3 | 22.2 | 23.0 |
| shp–es | 27.5 | 28.4 | 29.2 |

Table 11: The number of parameters (in millions) of our model.

| src-tgt | #Sents | OURS | | |
|---|---|---|---|---|
| | | Word | $+SW_{\text{ave}}$ | $+SW_{\text{cnn}}$ |
| nru–en | 605 | 2 | 3 | 7 |
| nru–zh | 1766 | 8 | 10 | 45 |
| nru–fr | 3833 | 21 | 31 | 212 |
| shp–es | 14276 | 96 | 160 | 1382 |
| ja–en | 331024 | 1565 | 1795 | – |

Table 12: Run-time (in seconds) of our model per one epoch on a single GPU.

### A.3 Run-time

Table 12 shows the run-time of our model in seconds. Although computationally more expensive than baseline models (except mBERT), it scales well and can be fully trained using a GPU in less than a minute for nru-zh.

## B Language Resources

Here we provide the URLs (in footnotes) from which we obtained the language resources we used in our experiments: Griko-Italian[37], Na-{English, French, Chinese}[38], Shipibo-Konibo-Spanish[39],

---

[37] https://bitbucket.org/antonis/grikoresource/src/master/ and https://github.com/antonisa/griko-italian-parallel-corpus
[38] https://github.com/alexis-michaud/na
[39] http://chana.inf.pucp.edu.pe/resources/parallel-corpus

| Source Word | Method | Retrieved Words |
|---|---|---|
| tʰi˧ (then, so, plane) | BIVEC | **then**, well, calculates, the, counts |
| | OURS ($+SW_{ave}$) | **then**, well, but, **so**, and |
| | + Pre | **then**, well, sikee, quickly, and |
| wɤ˧ (again) | BIVEC | also, **again**, place, roof, beams |
| | OURS ($+SW_{ave}$) | also, **again**, well, haenke, halves |
| | + Pre | also, haenke, **again**, and, once |
| ʑi˧mi˧ (house, farm) | BIVEC | **house**, building, houses, build, built |
| | OURS ($+SW_{ave}$) | **house**, building, main, room, build |
| | + Pre | **house**, building, home, houses, housed |
| tʂʰɯ˧-qo˧ (here) | BIVEC | **here**, fireplace, beam, halves, hearth |
| | OURS ($+SW_{ave}$) | **here**, sat, there, way, where |
| | + Pre | **here**, there, where, area, looking |
| mæ˩ (water) | BIVEC | story, edge, strong, told, eighteen |
| | OURS | cry, right, arrive, eat, strong |
| | + Pre | need, want, mupae, heard, terrifically |
| ə˧mi˧ (mother, aunt) | BIVEC | wow, **mother**, living, so, mmm |
| | OURS | wow, **mother**, alas, daughter, how |
| | + Pre | wow, **mother**, ggimi, truly, really |
| ɑ˩ʁo˧ (home) | BIVEC | **home**, family, lives, homes, else |
| | OURS ($+SW_{ave}$) | **home**, family, want, at, members |
| | + Pre | **home**, family, families, household, homes |

Table 13: Examples of retrieved words on nru-en BLI.

Japanese-English[40], English-Inuktitut[41], a Shipibo-Konibo-Spanish dictionary[42] and Na–French–Chinese–English dictionaries[43]. To download and pre-process the German-English and French-English data, we used the script at https://github.com/lilt/alignment-scripts, which is also used by Zenkel et al. (2020) and Chen et al. (2020). For other langauges, when the data is not pre-lowercased, we lowercased them. For ja-en, we applied SentencePiece to both languages independently and with the vocabulary size set to 20k. For en-iu, since there is a significant gap between the English and Inuktitut vocabulary size, i.e. 22k vs. 400k, we applied SentencePiece to Inuktitut only, with the vocabulary size set to 20k.

[40] http://www.phontron.com/kftt/
[41] http://web.eecs.umich.edu/~mihalcea/wpt05/
[42] http://chana.inf.pucp.edu.pe/resources/wordnet-shp
[43] https://github.com/alexis-michaud/na/tree/master/DICTIONARY

## C  Hyper Parameters of the Baselines

In this section, we describe the hyper parameters of the baselines we used in our experiments. For SENTID, we used the default settings since they are already optimised for low-resource data (25k sentences from Bible) and works well on both small and large data. For BIS2V, we set the number of negatives sampling to 5, max length of word n-gram to 2, and the number of n-gram dropout to 1. In high-resource experiments, we changed the number of n-gram dropout to 4. For BIVEC, we set the subsampling rate to 0.001, bi-weight to 2, and the number of negative sampling to 5. In high-resource experiments, we set the sampling value to 0.01 and the negative sampling value to 10. We observed BIVEC very sensitive to the subsampling rate and bi-weight: the performance ranged from 5.4 to 33.8 (P@1) in the nru-en BLI task. We tuned these hyperparameters of BIVEC and BIS2V based on the last-checkpoint model performance on the test data of nru-en and en-iu. For the mBERT baseline, we used their default setting.

For GIZA++ and Fast Align, we used their default hyper-parameters unless mentioned otherwise. In BLI, we set the counts increment cutoff and probability cutoff thresholds to 0 in GIZA++, and remove the probability cutoff in Fast Align to avoid pruning low-probability words.

## D Examples of Retrieved Words on BLI

Table 13 shows some examples of retrieved words on the Na-English BLI task. The source words are chosen by sorting the Na words in the dictionary based on the frequency in the Na-English corpus, and selecting the seven most frequent ones. The table shows that, although P@5 and P@1 are nearly the same for all the methods, our model matches more semantically and/or grammatically related words to the target word than BIVEC, the best performing baseline.[44] For instance, given the source word tʰi/, OURS ($+SW_{\text{ave}}$) retrieved its translations "then" and "so", and also other conjunctions "and" and "but", while BIVEC was able to retrieve "then" only and the other retrieved words are irrelevant to it, such as "calculates" and "counts". For the source word tsʰɯ˩-qo˩, all the models successfully retrieved its translation "here". However, while our models also retrieved relevant words to it, such as "there" and "where" , BIVEC retrieved completely irrelevant words such as "beam". These results suggest that our models encode more semantic and syntactic information into the word embeddings by taking word order information into account.

---

[44] Note that since the Na-English parallel corpus is extremely small, the size of the English vocabulary from which the words are retrieved is also very small (i.e. 942 word types). Besides, as shown in Table 2, the corpus is also very noisy and contains some ungrammatical sentences. Therefore, it is inevitable to some extent that some of the retrieved words are irrelevant to the source word.